\documentclass[lettersize,journal]{IEEEtran}
\usepackage{amsmath,amsfonts}
\usepackage{algorithmic}
\usepackage{algorithm}
\usepackage{array}
\usepackage[caption=false,font=normalsize,labelfont=sf,textfont=sf]{subfig}
\usepackage{textcomp}
\usepackage{stfloats}
\usepackage{url}
\usepackage{verbatim}
\usepackage{graphicx}
\usepackage{cite}
\usepackage{multirow}
\usepackage{xcolor}

\begin{document}

\title{Multi-Modal Character Localization and Extraction for Chinese Text Recognition}

\author{Qilong Li, Chongsheng Zhang
\thanks{This work was partially supported by the MOE Liberal Arts and Social Sciences Foundation (No.23YJAZH210), Major Program of National Social Science Foundation (No.23\&ZD309),  National Natural Science Foundation of China (No.62250410371), and Henan Provincial Center for Outstanding Overseas Scientists (No.GZS2025004). We thank the anonymous reviewers and the associate editor for their constructive suggestions and kind help. \textit{(Corresponding author: Chongsheng Zhang.)}}
\thanks{Qilong Li and Chongsheng Zhang are affiliated with the School of Computer and Information Engineering and the Henan Provincial Key Laboratory of Big Data Analysis and Processing, Henan University, Kaifeng, P.R. China. (E-mail: \{qilonghenu, cszhang\}@henu.edu.cn.)}
}



\maketitle

\begin{abstract}
Scene text recognition (STR) methods have demonstrated their excellent capability in English text images. However, due to the complex inner structures of Chinese and the extensive character categories, it poses challenges for recognizing Chinese text in images. Recently, studies have shown that the methods designed for English text recognition encounter an accuracy bottleneck when recognizing Chinese text images. This raises the question: Is it appropriate to apply the model developed for English to the Chinese STR task? To explore this issue, we propose a novel method named LER, which explicitly decouples each character and independently recognizes characters while taking into account the complex inner structures of Chinese. LER consists of three modules: Localization, Extraction, and Recognition. Firstly, the localization module utilizes multimodal information to determine the character's position precisely. Then, the extraction module dissociates all characters in parallel. Finally, the recognition module considers the unique inner structures of Chinese to provide the text prediction results. Extensive experiments conducted on large-scale Chinese benchmarks indicate that our method significantly outperforms existing methods. Furthermore, extensive experiments conducted on six English benchmarks and the Union14M benchmark show impressive results in English text recognition by LER. Code is available at \url{https://github.com/Pandarenlql/LER}. 

\end{abstract}


\section{Introduction}

\IEEEPARstart{S}{cene} text recognition (STR), which aims to read characters from the text images, is a fundamental task in computer vision and has attracted researchers' attention in the past years. Currently, a considerable number of methods have achieved favorable results in recognizing text in English instances. However, when these methods are directly applied to other languages, such as Chinese, satisfactory results are not obtained. 

Chinese, as a widely spoken language, is frequently encountered in daily life. Thus, recognizing Chinese characters from images is an urgent demand. Recently, studies have shown that directly applying the text recognition model developed for English images to Chinese images encounters an accuracy bottleneck \cite{yu2021benchmarking}. For instance, ABINet \cite{fang2021abinet} achieves an average recognition accuracy of merely 70.28\% in the Chinese scene. There might be two reasons: firstly, there are numerous character categories (exceeding 27,000); secondly, Chinese characters have a complex inner structure, and each character is composed of one or more radicals in accordance with spatial structures. Hence, developing a text recognition model that takes into account the unique characteristics of Chinese is a reasonable approach to breaking through the accuracy bottleneck. Currently, some researchers have made attempts to recognize Chinese characters from the perspective of character decomposition \cite{peng2022recognition, li2022tree, zhang2020radical, yang2021transformer, yu2023chinese, li2025ucr}. 

\begin{figure}[t]
  \centering
   \includegraphics[width=0.8\linewidth]{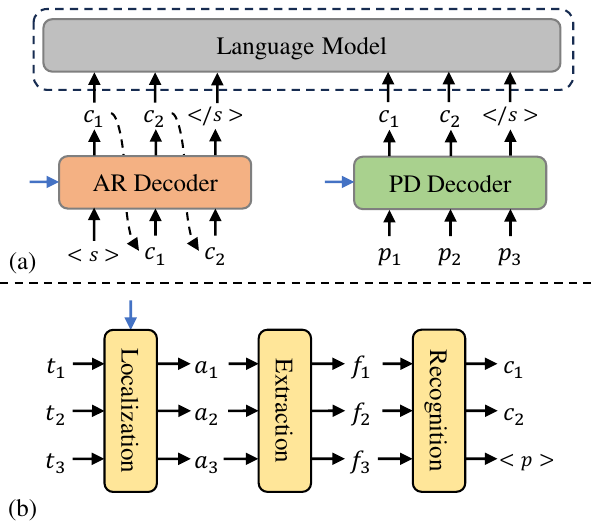}
   \caption{(a) Autoregressive decoder and parallel decoder. (b) The forward propagation of our LER. $<s>$, $</s>$, $<p>$, $C$, and $P$ denote the start of the sentence, the end of the sentence, pad, character, and position, respectively. $T$ denotes the text query proposed in this paper, $A$ denotes the visual feature obtained by localization module, $F$ denotes the independent character feature. 
   Capital letters (e.g., $T$) denote features, while lowercase letters (e.g., $t$) represent vectors within the features. Subscripts indicate the index of vectors (e.g., $t_i$).
   The blue flow denotes the visual feature obtained by the image encoder.}
   \label{fig:diff_arch}
\end{figure}

Previous text recognition methods can be summarized as autoregressive-based (AR) and parallel decoding-based (PD) according to the type of decoder, as shown in Fig. \ref{fig:diff_arch}(a).  
For AR methods, the visual features are first obtained using an image encoder, and then these features are decoded using an attention-based sequence decoder, such as an RNN \cite{graves2012long} or a Transformer \cite{vaswani2017attention}. However, in these methods, due to the decoder predicting characters step-by-step, an error in character prediction can make the subsequent recognition step unstable, leading to multiple character recognition errors, which is known as error accumulation.
To address this issue, some methods employ a language model for post-processing the predicted character sequence after AR decoding, leveraging the linguistic information of the sentence to correct erroneous characters \cite{wang2022petr, bhunia2021joint, na2022multi}. However, when the image itself contains writing errors, the language model may inadvertently correct characters based on linguistic information, potentially altering the original meaning of the text in the image. 
For PD methods, researchers have proposed utilizing position information as queries and employing cross-attention mechanisms to locate and decode characters from image features \cite{yu2020towards, fang2021abinet, wang2021two}. Compared with AR methods, this approach can recognize all characters simultaneously, resulting in higher efficiency. Additionally, since the query sequence is irrelevant to the text in the image, it avoids the error accumulation typically observed in AR methods. However, this method still has limitations. For instance, due to the high similarity between adjacent queries in the query sequence, attention drift is likely to occur \cite{zhang2023lpv}.
Our goal is to develop a text recognition model that is suitable for Chinese text recognition, while minimizing error accumulation and inadvertent corrections.

Explicit decoupling of each character in the image is an effective approach to avoid error accumulation. Therefore, we propose a novel text recognition framework that aims to achieve explicit decoupling of character features through two modules: multimodal character localization and extraction, as illustrated in Fig. \ref{fig:diff_arch}(b). First, the localization module utilizes CLIP text features and linguistic information to locate each character within the visual features accurately. Then, the extraction module explicitly decouples the features of each character. Finally, the recognition module predicts each character feature independently. Since each character feature is processed independently during recognition, this approach fundamentally avoids error accumulation. Additionally, this framework also avoids the issue of automatic correction due to lexical dependence when recognizing meaningless text or multi-word text \cite{jiang2023revisiting}.

This paper introduces a multimodal parallel decoding framework named LER for Chinese text recognition, which consists of three modules: Localization, Extraction, and Recognition, as shown in Fig. \ref{fig:methods}.
Specifically, the localization module is a multimodal model designed to accurately locate characters within images by leveraging CLIP text features and linguistic information, thereby enhancing localization accuracy and mitigating attention drift. The extraction module focuses on extracting and constructing visual features for each character's region based on the localization results. The recognition module is designed to recognize decoupled characters. 
During training, we employ an IDS decoder to recognize the radical sequences of decoupled characters. This enhances the performance of the extraction module, as a decoupled character can only be correctly recognized if it contains complete radicals.
Notably, the IDS decoder is utilized exclusively during training and can be omitted during testing to reduce the model's parameter number and enhance computational efficiency.
Importantly, unlike most text recognition models that rely on implicit character encoding, our method explicitly extracts and decouples character features in the extraction module, enabling independent character recognition and avoiding error accumulation and automatic error correction.

The main contributions of our work are as follows: 

\begin{itemize}
    \item We propose a novel parallel decoding multimodal framework, LER, for accurate Chinese text recognition.

    \item We propose a novel multimodal character localization method that leverages the image-text alignment capabilities of CLIP and linguistic information to locate characters precisely. 

    \item We propose a character extraction method that achieves explicit decoupling of individual characters, enabling the model to independently recognize each character and thereby avoid error accumulation and automatic error correction.

    \item LER achieves state-of-the-art performance on Chinese text recognition benchmarks with a relatively small model size. Additionally, experiments demonstrate that our method also performs well on English images.
\end{itemize}

\begin{figure}[t]
  \centering
   \includegraphics[width=0.7\linewidth]{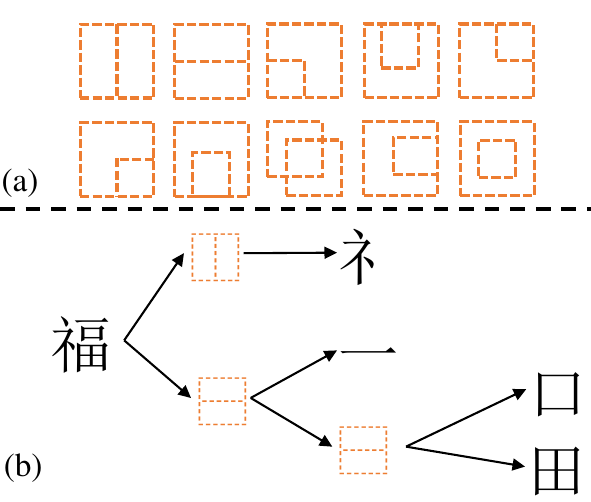}
   \caption{(a) The spatial structure of Chinese characters. (b) Illustration of character decomposition.}
   \label{fig:ids}
\end{figure}

\section{Preliminaries}
\subsection{Radical-based Chinese Character Decomposition}
According to the official Chinese standard GB 18030-2000 \footnote{https://zh.wikipedia.org/wiki/GB\_18030}, there are a total of 27,533 characters, with 3,755 classified as common first-level characters. The vast number of character types poses significant challenges for Chinese text recognition. Fortunately, several decomposition methods exist that allow all Chinese characters to be systematically decomposed into sequences of basic components and spatial structures, such as the IDS (Ideographic Description Sequence) decomposition method. 
IDS is defined in the international ISO/IEC 10646 standard, as illustrated in Fig. \ref{fig:ids}, it is a hierarchical tree-based method that incorporates 10 spatial structures and 562 radicals.  

\subsection{Related Work}

\begin{figure*}[t]
  \centering
  \includegraphics[width=1.0\linewidth]{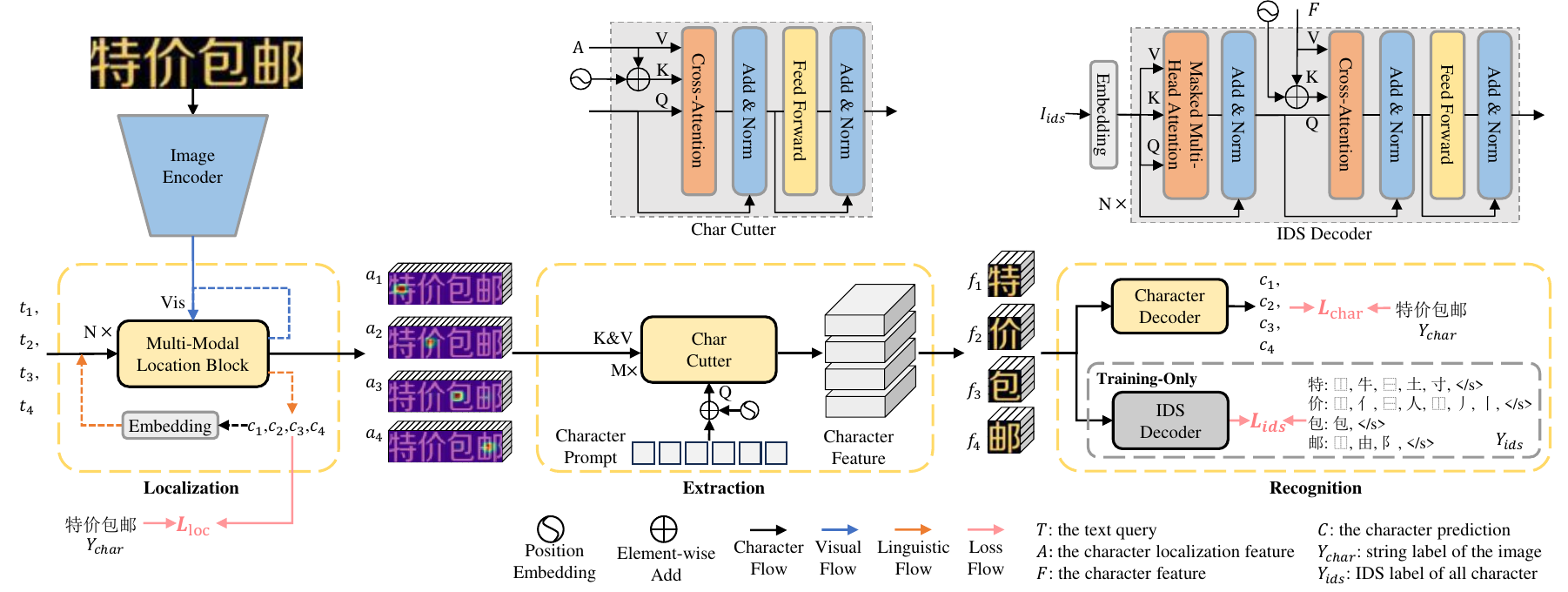}
  \caption{The proposed LER framework and IDS decoder, $N$ and $M$ denote the number of blocks in the LER network and the number of char cutter blocks in the extraction module, respectively. }
  \label{fig:methods}
\end{figure*}

Scene text recognition (STR) aims to transcribe text-line images into character sequences. CRNN \cite{shi2016end} proposed the CNN-RNN hybrid architecture with CTC decoding, while SVTR \cite{du2022svtr} introduced novel mixing mechanisms in the visual encoder to enhance CTC-based recognition.

Recent STR methods primarily adopt either autoregressive (AR) or parallel decoding (PD) frameworks. The AR paradigm originated from attention-based RNN decoders \cite{graves2012long, chung2014empirical, zhang2018track, shi2018aster, li2019show, zhang2020srd, wang2020decoupled}, later evolving into transformer decoder \cite{vaswani2017attention, sheng2019nrtr, bautista2022scene, xie2022toward}. Furthermore, TrOCR \cite{li2023trocr} adopts a complete transformer framework for text recognition, achieving state-of-the-art accuracy. 
LevOCR \cite{da2022levenshtein} uniquely combines visual recognition with iterative language model refinement. 

PD methods address speed limitations through simultaneous decoding. Researchers have proposed the use of positional encoding as queries to simultaneously locate and recognize text \cite{fang2021abinet, wang2021two}.
Zhang et al. \cite{zhang2023lpv} resolved attention drift via cascaded recognition modules, and PIMNet \cite{qiao2021pimnet} introduced iterative parallel decoding. 

In recent years, the advancement of multimodal models has revealed significant potential across various downstream tasks. 
Notably, the large-scale Contrastive Language-Image Pretraining (CLIP) model \cite{radford2021learning} has demonstrated significant effectiveness in achieving image-text alignment.
In scene text detection and recognition, several studies have explored the use of CLIP to improve detection and recognition accuracy \cite{yu2023chinese, yu2023turning}. 
Given CLIP's superior capability in aligning multimodal information, it may offer a promising approach to addressing the attention drift in parallel decoding STR methods.

\section{Methodology}
\subsection{Overview}

LER's pipeline (Fig. \ref{fig:methods}) comprises: (1) CNN image encoder converting input ($H \times W \times 3$) to visual features $Vis \in R^{\frac{H}{4} \times \frac{W}{4} \times D_0}$; (2) Multimodal localization module generating character location features $A \in R^{L \times \frac{H}{4} \times \frac{W}{4} \times D_1}$ ($L$: max text length); (3) Extraction module decoding $A$ into character features $F \in R^{L \times c_h \times c_w \times D_2}$ ($c_h/c_w$: character dimensions); (4) Recognition module predicting from $F$.
The IDS decoder is exclusively used during the training stage, improving the accuracy of radical-aware Chinese character recognition and segmentation. During inference, only the character decoder is active.

\subsection{Localization Module}

Precise character localization is crucial for robust text recognition. While parallel decoding methods, such as ABINet \cite{fang2021abinet}, employ positional queries for character detection, their sequential similarity often induces attention drift \cite{zhang2023lpv}. To address this challenge, we propose the Multimodal Localization Block (MLB), which dynamically integrates CLIP's cross-modal alignment capabilities with linguistic context through an N-stage cascaded architecture. Each MLB block processes visual features through progressively refined text embeddings $T^i_j$ (where $i \in \{1,...,N\}$ denotes the block index and $j \in \{1,...,L\}$ indicates character position). The initial stage ($T^1$) bootstraps localization using CLIP's pretrained text features, while subsequent iterations incorporate the previous MLB's predicted embeddings. This hierarchical refinement produces increasingly discriminative queries that simultaneously encode visual-textual correlations and linguistic patterns, effectively suppressing attention drift through multi-modal feature fusion. The resulting localization module achieves sub-character precision by synergistically combining visual evidence with contextual language constraints.

\begin{figure}[t]
  \centering
   \includegraphics[width=1.0\linewidth]{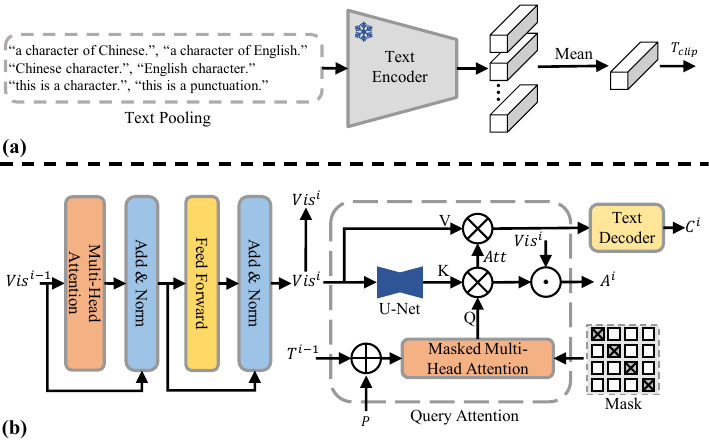}
   \caption{Multimodal Localization Block. (a) The CLIP's text feature. (b) The structure of the Multimodal Localization Block (MLB). $Vis$, $T$, $P$, $A$, and $C$ denote the visual feature, the text query, the position embedding, the localization feature, and character, respectively. Superscript denotes the index of MLB in the localization module.}
   \label{fig:mlb}
\end{figure}

In the first iteration of the localization module, $T^1$ utilizes CLIP's text features to assist in character localization and recognition, it contains two parts: a content-free text query $T_{clip}$ and position query $P$.
Inspired by TCM \cite{yu2023turning}, we adopt the pre-trained CLIP text encoder while freezing its parameters. 
Leveraging the text-image alignment capability of CLIP, we propose using simple, content-agnostic sentences as prompts, such as ``a character of Chinese,'' as queries for locating characters in images.
These features from CLIP's text encoder can effectively highlight regions containing Chinese characters within the image. 
However, crafting optimal text prompts remains challenging. To address this, we introduce a flexible method for generating text queries. Specifically, we first establish a text pool $Text=\{s_0, s_1, ..., s_k\}$ comprising sentences with diverse linguistic contents, where each $s_i$ is an independent sentence and $k$ denotes the total number of sentences. Then, we encode all sentences using the CLIP text encoder to obtain their respective features. Finally, we aggregate these features to derive a semantically coherent representation $T_{clip}$. 
All content-agnostic sentences are shown in Fig. \ref{fig:mlb} (a), which are used to enable character location capability via linguistic prompt. The formulation for computing the text query is as follows:
\begin{equation}
\begin{split}
    T_t'&=TextEncoder(Text) \in R^{k \times D_0} \\
    T_{clip} &= Mean(T_t') \in R^{D_0}                 \\
\end{split}
\end{equation}

In each iteration of the localization module, the position encoding is added with $T^i$; therefore, the first query $T^1$ is as follows:
\begin{equation}
\begin{split}
    T^1=T_{clip} + P \in R^{L \times D_0}
\end{split}
\end{equation}

The proposed MLB contains two components, as shown in Fig. \ref{fig:mlb} (b). The first is a transformer encoder block, which aims to enhance visual features. The second component is the query attention mechanism, designed to locate characters accurately. 
Specifically, the query attention builds upon and improves the position attention mechanism \cite{fang2021abinet}, enabling more precise character localization. 
Query attention incorporates an additional Masked Multi-head Self-Attention (MMSA) layer before computing cross-attention compared with position attention, with a diagonal mask applied in the MMSA layer.
This is because, after the second MLB iteration, the input $T^i$ contains linguistic information. By leveraging this linguistic information through the MMSA layer, the model can precisely locate the character.
The used diagonal mask is:
\begin{equation}
\begin{split}
    M_{ij} &= \left\{\begin{matrix}
                      0, & i\ne j\\
                      -\infty, & i = j
                    \end{matrix}\right. \\
\end{split}
\label{eq:mmsa}
\end{equation}

Subsequently, the character attention map $Att \in R^{L \times \frac{HW}{16}}$ , the prediction results of characters in the image $C^i \in R^{L \times n_{class}}$ and the visual feature map $A^i \in R^{L \times \frac{HW}{16} \times D_0}$ are computed according to Eq. \ref{eq:MLB}.
\begin{equation}
\begin{split}
    Att &= softmax(\frac{KQ^T}{\sqrt{D_0}}) \\
    C^i &= TD(Att \times V)   \\
    A^i &= Att \cdot Vis^i  \\
\end{split}
\label{eq:MLB}
\end{equation}
where $TD$ is the text decoder that predicts the character category.

After the localization module, the final visual feature $A=\{a_1, a_2, ..., a_j\}, j \in L$ is obtained. Each character feature map $A_j$ represents the feature of the $j$-th character, as illustrated in Fig. \ref{fig:methods}. 
Due to the attention mechanism focusing exclusively on the region occupied by each character, only the features corresponding to the $j$-th character's region are retained in $A_j$, while features from other regions are suppressed. Subsequently, $A$ is fed into the extraction module for the distinct decoupling of each character.

\subsection{Extraction Module}
The extraction module aims to isolate character features from the visual features generated by the localization module. Given that each feature map in $A \in R^{L \times \frac{HW}{16} \times D_1}$ contains a significant amount of redundant information, it is essential to extract relatively independent and meaningful character features. To this end, this paper introduces a char cutter within this module to achieve this objective.

The structure of the char cutter is illustrated in Fig. \ref{fig:methods}. 
In the char cutter, first, the height $c_h$ and width $c_w$ of the decoupled character need to be defined. 
Then, a char prompt of length $c_l = c_h \times c_w$ is constructed as the input of the char cutter. 
Finally, through $M$ cross-attention layers, the decoupled character features $F \in R^{L \times c_l \times D_1}$ are obtained. After reshaping, the final character features are represented as $F \in R^{L \times c_h \times c_w \times D_1}$. In the cross-attention layers, the key ($K$) and value ($V$) are derived from the output of the localization module ($A$), as the feature $A$ inherently contains character localization information, which facilitates precise extraction of character features.

Owing to the relative independence of character features $F$ obtained by the extraction module, the recognition for each character operates independently. Consequently, this independence ensures that recognition results do not suffer from error accumulation and automatic error correction.

\subsection{Recognition Module}
Given the complex inner structures of Chinese characters, considering their unique radical decomposition process during training could potentially enhance recognition performance. 

In the recognition module, we present two decoders: the character decoder and the IDS decoder. 
Both decoders receive the character features $F$ as input and predict the category and IDS, respectively.
The character decoder is a classifier. 
Specifically, the character feature $F$ is aggregated by a global pooling layer, then a linear layer is employed to obtain the character category feature $C \in R^{L \times n_{class}}$, where $n_{class}$ denotes the number of supported character categories.

The IDS decoder is employed to predict the radical sequence of the decoupled character feature $F$. As illustrated in the Fig. \ref{fig:methods}, this IDS decoder comprises $N$ transformer decoder blocks. In the cross-attention layer, both key and value are derived from character features $F$. The IDS decoder outputs IDS prediction features $C_{ids} \in R^{L \times L_{ids} \times n_{ids}}$ for each character, where $L_{ids}$ denotes the max length of the IDS sequence and $n_{ids}$ represents the number of radical categories.

Moreover, since the IDS decoder can identify whether all radicals constituting a character are present in the character features $F$, its integration aids the extraction module in improving the capacity of the char cutter, ensuring that the extracted character features are as comprehensive as possible.

\subsection{Training Objective}
During the training process, LER integrates three distinct loss functions: character localization loss $L_{loc}$, character recognition loss $L_{char}$, and IDS prediction loss $L_{ids}$. Each of these losses is formulated as a cross-entropy loss, but they serve different objectives. Specifically, $L_{loc}$ is employed in the localization module to ensure precise character localization. $L_{char}$ is utilized to train the LER for accurate character recognition. Meanwhile, $L_{ids}$ aims to ensure precise cropping of characters, thereby enabling correct identification of all radicals. The overall localization loss is computed across $N$ iterations of prediction results, where the localization stage prediction is defined as $C_{loc}=\{C^0_{loc}, C^1_{loc},..., C^i_{loc}\}, i \in N, C_{loc} \in R^{L \times n_{class}}$. The character prediction of the char decoder in the recognition module is defined as $C_{char}$
The LER gives predictions for $L$ characters in parallel. In labels, the text lengths shorter than $L$ are padded with $<p>$ to ensure a uniform length of $L$.
The formulation of each loss function is detailed as:
\begin{equation}
\begin{split}
    &L_{loc}=\sum_{i=1}^{N} CE(C^i_{loc}, Y_{char})\\
    &L_{char}=CE(C_{char}, Y_{char})\\
    &L_{ids}=CE(C_{ids}, Y_{ids})\\
    &L = L_{char} + L_{ids} + L_{loc}\\
\end{split}    
\label{eq:loss}
\end{equation}
where $CE$ is the cross-entropy loss, $Y_{char}$ is the text label, $Y_{ids}$ is the IDS label.

\subsection{Training Strategy}
\label{subsec:training_strategy}
During the early training stage, the localization module struggles to accurately determine character positions, which hinders the effective training of the extraction and recognition modules. 
Consequently, we design a two-stage training strategy for the LER network. In the first stage, we focus on training the image encoder and localization module using the localization loss function $L_{loc}$. In the second stage, we apply the overall loss function to train all components of the network jointly. 
While the network used in the first stage possesses text recognition capabilities, its reliance on linguistic information during text prediction may introduce error accumulation and automatic error correction. Therefore, the primary objective of the first stage is to enhance character localization accuracy through multimodal information fusion, rather than achieving optimal text recognition accuracy.

\begin{table}[t]
\centering
\caption{Architecture specifications of LER. TDB denotes the transformer decoder block.}
\resizebox{\columnwidth}{!}{%
\begin{tabular}{cccccc}
\hline
Model & Operator & Out Size & Depth & Head & Dim \\
\hline
Image Encoder & Conv Mix & (H/4,W/4) & 6 & - & 128 \\
Localization & MLB & (H/4,W/4) & 6 & 4 & 128 \\
Extraction & Char cutter & ($c_h$, $c_w$) & 3 & 8 & 256 \\
IDS Decoder & TDB & ($L_{ids}$, $n_{ids}$) & 6 & 8 & 384\\
\hline
\end{tabular}%
}
\label{tab:LER_details}
\end{table}

\section{Experiments}
\subsection{Datasets}
We validate LER on both Chinese and English datasets. For Chinese, we use the Chinese Text Recognition (CTR) benchmark \cite{yu2021benchmarking}, which contains four scenarios of Chinese text images: scene, web, document, and handwriting. Each scenario includes training, validation, and test datasets. 
Following the settings in \cite{du2024smtr}, we combine the training datasets of all four scenarios in the training stage, and in the test stage, we evaluate the model on each respective test dataset.

For English, we use Union14M-L dataset \cite{jiang2023revisiting},  which aggregates images from a majority of existing public text recognition datasets, covering a broad spectrum of real-world scenarios.
In the English experiments, we train LER on the Union14M-L training dataset and evaluate its performance on both the Union14M benchmark and six commonly used benchmarks: IIIT \cite{mishra2012scene}, IC13 \cite{karatzas2013icdar}, SVT \cite{wang2011end}, IC15 \cite{karatzas2015icdar}, SVTP \cite{phan2013recognizing}, and CUTE \cite{risnumawan2014robust}.

\subsection{Implementation Details}

Inspired by the lightweight text recognition network SVTR \cite{du2022svtr}, we developed a lightweight LER network using its conv mix encoder and merging operations. As shown in Table \ref{tab:LER_details}.

Following section \ref{subsec:training_strategy}, we use the AdamW optimizer ($\lambda=0.05$) \cite{loshchilov2018adamw} to train LER and the rectification module to correct the distorted text images\cite{shi2018aster}. For Chinese, the input images are without augmentation and have a size of $32\times320$ after rectification, the cut character size is $(c_h,c_w)=(4,4)$. In contrast, English uses $32\times100$ images with augmentation, the cut character size is $(c_h,c_w)=(4,2)$. Both employ cosine schedulers (Chinese: 100 epochs for stage 1 and 30 epochs for stage 2, 5 warm-up epochs; English: 20 epochs for stage 1 and 15 epochs for stage 2, 4 warm-up epochs) at an initial learning rate of $3e^{-4}$, the maximum sequence length $L=25$. All experiments are trained on four NVIDIA RTX 4090 GPUs. 

\subsection{Evaluation Metrics}
Following the previous CTR works \cite{yu2023chinese}, we select Line-level text recognition ACCuracy (LACC) and Normalized Edit Distance (NED) to evaluate our method. LACC is defined as:
\begin{equation}
    LACC=\frac{1}{N} \sum_{N}^{1} \mathbb{I}(C=Y_{char})
\end{equation}
where $N$ is the number of test images, $\mathbb{I}$ is the indicator function. NED is defined as:
\begin{equation}
    NED= 1 - \frac{1}{N} \sum_{N}^{1} ED(C, Y_{char})/Maxlen(C, Y_{char})
\end{equation}
where $ED$ is edit distance, $Maxlen$ is the maximum sequence length. 
NED is an appropriate metric to evaluate error accumulation because it quantifies the proportion of error characters relative to the total number of characters in a prediction text. For instance, consider a specific text image where one prediction, denoted as $C_1$, contains one error character, while another prediction, denoted as $C_2$, contains two error characters. Although both $C_1$ and $C_2$ have an LACC (Line Accuracy) of 0, the NED of $C_1$ is larger than that of $C_2$, indicating that $C_1$ is more robust overall. 

\subsection{Experiments on Chinese Benchmark}
To evaluate the performance of LER in Chinese text recognition, we conducted a comparative experiment with several excellent text recognition methods. Except for SMTR, all methods are trained exclusively on their respective scenario-specific training datasets and subsequently tested in corresponding scenarios. In contrast, both SMTR and LER are trained using the whole training dataset. 
The IDS decoder is excluded when we calculate the LER parameters. As illustrated in Table \ref{tab:ctr}, LER consistently outperformed other models across all scenarios, achieving superior recognition accuracy while maintaining a relatively smaller number of parameters. 
Although LER's average accuracy improvement over the current state-of-the-art model, SMTR, is 1.07\%, it demonstrates a significant 2.4\% accuracy improvement in the complex handwritten scenario, which is more challenging due to dense text and a variety of writing styles.

\begin{table*}[th]
\centering
\caption{Comparison with previous methods on the CTR benchmark. The metrics are LACC / NED.}
\begin{tabular}{ccccccc}
\hline
\multirow{2}{*}{Method} & \multicolumn{4}{c}{Dataset} & \multirow{2}{*}{Average(\%)} & \multirow{2}{*}{Parameters(M)} \\ \cline{2-5}
 & Scene & Web & Document & Handwriting &  &  \\ \hline
CRNN\cite{shi2016end} & 53.41 / 0.712 & 57.00 / 0.716 & 96.62 / 0.992 & 50.83 / 0.814 & 64.47 / 0.809 & 12.4 \\
ASTER\cite{shi2018aster} & 61.34 / 0.815 & 51.67 / 0.715 & 96.19 / 0.991 & 37.00 / 0.683 & 61.55 / 0.801 & 27.2 \\
MORAN\cite{luo2019moran} & 54.61 / 0.684 & 31.47 / 0.446 & 86.10 / 0.962 & 16.24 / 0.305 & 47.10 / 0.559 & 28.5 \\
SAR\cite{li2019show} & 59.67 / 0.766 & 58.03 / 0.716 & 95.67 / 0.988 & 36.49 / 0.736 & 62.48 / 0.802 & 27.8 \\
SEED\cite{qiao2020seed} & 44.72 / 0.681 & 28.06 / 0.460 & 91.38 / 0.980 & 20.97 / 0.475 & 46.30 / 0.549 & 36.1 \\
MASTER\cite{lu2021master} & 62.82 / 0.726 & 52.05 / 0.620 & 84.39 / 0.944 & 26.92 / 0.443 & 56.55 / 0.683 & 62.8 \\
ABINet\cite{fang2021abinet} & 66.55 / 0.792 & 63.17 / 0.776 & 98.19 / 0.996 & 53.09 / 0.813 & 70.28 / 0.844 & 53.1 \\
TransOCR\cite{chen2021scene} & 71.60 / 0.834 & 65.52 / 0.782 & 97.36 / 0.994 & 53.67 / 0.802 & 71.55 / 0.853 & 83.9 \\
RTN\cite{yang2021transformer} & 70.41 / 0.827 & 65.70 / 0.793 & 97.23 / 0.994 & 55.96 / 0.806 & 72.33 / 0.855 & 74.2 \\
SAN\cite{zhang2023san} & 73.6 / 0.863 & 67.3 / 0.817 & -- / -- & -- / -- & -- & -- \\
CCR-CLIP\cite{yu2023chinese} & 71.31 / 0.829 & 69.21 / 0.797 & 98.29 / 0.997 & 60.30 / 0.849 & 74.78 / 0.868 & 62.0 \\
PARSeq\cite{bautista2022scene} &  74.3 / -- & 74.6 / -- & 96.9 / -- & 55.9 / -- & 76.10 / -- & -- \\
CPPD\cite{du2023context} & 74.4 / -- & 76.1 / -- & 98.6 / -- & 55.3 / -- & 76.10 / -- & 29.4 \\
SMTR\cite{du2024smtr} & 79.8 / -- & 80.6 / -- & 99.1 / -- & 61.9 / -- & 80.33 / -- & 20.8 \\ \hline
LER &  \textbf{81.36 / 0.906} & \textbf{80.81 / 0.902} & \textbf{99.33 / 0.999} & \textbf{64.38 / 0.910} & \textbf{81.47 / 0.929} & 25.2 \\ \hline
\end{tabular}%
\label{tab:ctr}
\end{table*}

Additionally, Section \ref{subsec:ablation} presents further experiments where LER is trained solely on the handwritten scenario training dataset and tested on the corresponding test dataset, following the same experimental settings as other methods, except for SMTR. The results show that LER achieved the highest performance in both LACC and NED. 
Specifically, the improved NED indicates that LER's explicit decoupling and independent character recognition mechanism yield a robust model.

\begin{table*}[th]
\centering
\caption{Results on the six common English benchmarks, \textbf{bond} and \underline{underline} values denote the 1st and 2nd results in each column.}
\begin{tabular}{cc|ccccccc|c}
\hline
Type & Method & IIIT & IC13 & SVT & IC15 & SVTP & CUTE & Avg & Paramters(M) \\ \hline
\multirow{2}{*}{CTC} & CRNN\cite{shi2016end} & 90.8 & 91.8 & 83.8 & 71.8 & 70.4 & 80.9 & 81.6 & 8.3 \\
 & SVTR\cite{du2022svtr} & 95.9 & 95.5 & 92.4 & 83.9 & 85.7 & 93.1 & 91.1 & 24.6 \\ \hline
\multirow{5}{*}{AR} & MORAN\cite{luo2019moran} & 96.7 & 94.6 & 91.7 & 84.6 & 85.7 & 90.3 & 90.6 & 17.4 \\
 & ASTER\cite{shi2018aster} & 96.1 & 94.9 & 93.0 & 86.1 & 87.9 & 92.0 & 91.7 & 19.0 \\
 & DAN\cite{wang2020decoupled} & 97.5 & 96.5 & 94.7 & 87.1 & 89.1 & 94.4 & 93.2 & 27.7 \\
 & RobustScanner\cite{yue2020robustscanner} & 98.5 & 97.7 & 95.8 & 88.2 & 90.1 & \underline{97.6} & 94.7 & 48.0 \\
 & SATRN\cite{lee2020recognizing} & 97.0 & 97.9 & 95.2 & 87.1 & 91.0 & 96.2 & 93.9 & --\\
 & BUSNet\cite{wei2024image} & 98.3 & 97.8 & \textbf{98.1} & \textbf{90.2} & \underline{95.3} & 96.5 & \underline{96.1} & 32.1 \\
 & CDistNet\cite{zheng2024cdistnet} & 98.7 & 97.8 & 97.1 & 89.6 & 93.5 & 96.9 & 95.6 & 43.3 \\
 & OTE-AR\cite{xu2024ote} & 98.1 & \underline{98.0} & 98.0 & 89.1 & \textbf{95.5} & \underline{97.6} & \underline{96.1} & 25.2 \\
 & PARSeq\cite{bautista2022scene} & \textbf{98.9} & \textbf{98.4} & \textbf{98.1} & \underline{90.1} & 94.3 & \textbf{98.6} & \textbf{96.4} & 23.8 \\ \hline
\multirow{5}{*}{PD} & SRN \cite{yu2020towards} & 95.5 & 94.7 & 89.5 & 79.1 & 83.9 & 91.3 & 89.0 & 55 \\
 & VisionLAN\cite{wang2021two} & 96.3 & 95.1 & 91.3 & 83.6 & 85.4 & 92.4 & 91.3 & 32.8 \\
 & ABINet\cite{fang2021abinet} & 97.2 & 97.2 & 95.7 & 87.6 & 92.1 & 94.4 & 94.0 & 36.7 \\
 & LPV-B\cite{zhang2023lpv} & \textbf{98.9} & 97.4 & \underline{97.4} & 89.8 & 93.0 & 97.2 & 95.6 & 30.5 \\
 & OTE\_PD\cite{xu2024ote} & 98.1 & 97.5 & 96.6 & 86.7 & 91.2 & 96.2 & 93.4 & 25.2 \\ \hline
 & LER & \underline{98.8} & 97.8 & 97.3 & 89.8 & 93.2 & \textbf{98.6} & 95.9 & 19.7 \\ \hline
\end{tabular}%
\label{tab:commonbenchmark}
\end{table*}

\subsection{Experiments on English Benchmark}
To evaluate the performance of LER in English text recognition, we conduct experiments that compare with 14 STR methods on multiple English benchmark datasets. These methods are categorized based on their decoder types into three groups: CTC decoders, AR decoders, and PD decoders. We trained the LER model on the Union14M-L training dataset and evaluated its performance on six common benchmarks as detailed in Table \ref{tab:commonbenchmark}. Despite using fewer parameters, LER demonstrated performance comparable to PARSeq and OTE-AR, thereby validating its effectiveness.

\begin{table*}[ht]
\centering
\caption{Results on the Union14M benchmarks, \textbf{bond} and \underline{underline} values denote the 1st and 2nd results in each column.}
\begin{tabular}{cc|cccccccc|c}
\hline
\multirow{2}{*}{Type} & \multirow{2}{*}{Method} & Curve & Multi- & Artistic & Context & Salient & Multi- & General & \multirow{2}{*}{Avg} & \multirow{2}{*}{Parameters(M)} \\
 &  &  & Oriented &  & -less &  & Words &  &  &  \\ \hline
\multirow{2}{*}{CTC} & CRNN\cite{shi2016end} & 19.4 & 4.5 & 34.2 & 44.0 & 16.7 & 35.7 & 60.4 & 30.7 & 8.3 \\
 & SVTR\cite{du2022svtr} & 72.4 & 68.2 & 54.1 & 68.0 & 71.4 & 67.7 & 77.0 & 68.4 & 24.6 \\ \hline
\multirow{5}{*}{AR} & MORAN\cite{luo2019moran} & 51.2 & 15.5 & 51.3 & 61.2 & 43.2 & 64.1 & 69.3 & 50.8 & 17.4 \\
 & ASTER\cite{shi2018aster} & 70.9 & 82.2 & 56.7 & 62.9 & 73.9 & 58.5 & 76.3 & 68.8 & 19.0 \\
 & DAN\cite{wang2020decoupled} & 74.9 & 63.3 & 63.4 & 70.6 & 70.2 & 71.1 & 76.8 & 70.1 & 27.7 \\
 & RobustScanner\cite{yue2020robustscanner} & 79.4 & 68.1 & 70.5 & \underline{79.6} & 71.6 & \textbf{82.5} & 80.8 & 76.1 & 48.0 \\
 & SATRN\cite{lee2020recognizing} & 74.8 & 64.7 & 67.1 & 76.1 & 72.2 & 74.1 & 75.8 & 72.1 & -- \\
 & BUSNet\cite{wei2024image} & 83.0 & \underline{82.3} & 70.8 & 77.9 & 78.8 & 71.2 & \underline{82.6} & 78.1 & 32.1 \\
 & CDistNet\cite{zheng2024cdistnet} & 81.7 & 77.1 & 72.6 & 78.2 & \textbf{79.9} & \underline{79.7} & 81.1 & \textbf{78.6} & 43.3 \\
 & OTE-AR\cite{xu2024ote} & \underline{83.1} & \textbf{82.8} & \underline{73.5} & 73.7 & \underline{79.7} & 70.3 & 82.2 & 77.9 & 25.2 \\ \hline
\multirow{5}{*}{PD} & SRN \cite{yu2020towards} & 49.7 & 20.0 & 50.7 & 61.0 & 43.9 & 51.5 & 62.7 & 48.5 & 55 \\
 & VisionLAN\cite{wang2021two} & 70.7 & 57.2 & 56.7 & 63.8 & 67.6 & 47.3 & 74.2 & 62.5 & 32.8 \\
 & ABINet\cite{fang2021abinet} & 75.0 & 61.5 & 65.3 & 71.1 & 72.9 & 59.1 & 79.4 & 69.2 & 36.7 \\
 & LPV-B\cite{zhang2023lpv} & 82.4 & 64.6 & \textbf{74.1} & \textbf{81.0} & 78.8 & 81.1 & \textbf{82.8} & 77.8 & 30.5 \\
 & OTE\_PD\cite{xu2024ote} & 79.2 & 76.0 & 70.0 & 74.3 & 76.0 & 64.2 & 80.1 & 74.3 & 25.2 \\ \hline
 & LER & \textbf{83.3} & 80.2 & 71.2 & 76.8 & 76.6 & 78.5 & 81.5 & \underline{78.3} & 19.7 \\ \hline
\end{tabular}%
\label{tab:unionbenchmark}
\end{table*}

Furthermore, we extended our experiments by testing the LER model on the Union14M Benchmarks, as presented in Table \ref{tab:unionbenchmark}. Consistent with the results from the common benchmarks, LER achieved competitive performances while maintaining a relatively small model size. Notably, methods relying on AR decoders and language model post-processing exhibit significant limitations in ``Contextless'' scenarios due to lexicon dependencies developed during training, which can lead to automatic error correction \cite{jiang2023revisiting}. 
In contrast, the LER model recognizes each character independently, making it more robust in these challenging scenarios compared to AR-based methods with similar model sizes.

As demonstrated in Table \ref{tab:commonbenchmark} and Table \ref{tab:unionbenchmark}, the LER model exhibits slightly inferior performance compared to state-of-the-art methods on English benchmarks. This phenomenon can be primarily attributed to the fundamental differences in character structures between English and Chinese, which render the IDS decoder redundant when training English text images. Nevertheless, LER achieves competitive accuracy relative to state-of-the-art models while utilizing substantially fewer parameters. Our ablation studies further confirm that enlarging the model architecture (e.g., scaling up to LER-L) yields measurable improvements in accuracy, demonstrating LER's potential in high-performance applications.

\subsection{Ablation Experiments}
\label{subsec:ablation}
To better understand LER, we conduct experiments to validate the proposed modules.

\begin{table}[t]
\centering
\caption{Effectiveness of each module in the LER network. $\checkmark^-$ means without CLIP features in the localization module.}
\renewcommand{\arraystretch}{1.15}
\begin{tabular}{cccccc}
\hline
\multirow{2}{*}{Localization} & \multirow{2}{*}{Extraction} & \multicolumn{2}{c}{Recognition} & \multirow{2}{*}{Handwriting} & \multirow{2}{*}{Time (ms)} \\ \cline{3-4}
 &  & Char & \multicolumn{1}{c}{IDS} &  &  \\ \hline
$\checkmark^-$ &  &  &  & 56.53 / 0.847 & \multicolumn{1}{c}{6.1} \\
$\checkmark$ &  &  &  & 57.22 / 0.850 & \multicolumn{1}{c}{6.1} \\
$\checkmark$ & \multicolumn{1}{l}{} & $\checkmark$ &  & \multicolumn{1}{l}{57.34 / 0.851} & \multicolumn{1}{c}{6.3} \\
$\checkmark$ & $\checkmark$ & $\checkmark$ &  & 57.75 / 0.855 & \multirow{2}{*}{6.9} \\
$\checkmark$ & $\checkmark$ & $\checkmark$ & \multicolumn{1}{c}{$\checkmark$} & 59.29 / 0.871 &  \\ \hline
\end{tabular}%
\label{tab:ablation_modules}
\end{table}

\textbf{Effectiveness of each module.} In this experiment, we aim to prove the effectiveness of the proposed modules. 
We systematically validated the effectiveness of the proposed modules through three metrics: LACC, NED, and inference time, all experiments are conducted on an NVIDIA RTX 4090 GPU. 
Furthermore, we designed an ablation experiment on CLIP features within the localization module. 
To verify the efficacy of the extraction module, we directly fed the output of the localization module into the character decoder and compared its accuracy with that achieved after incorporating the extraction module.
As illustrated in Table \ref{tab:ablation_modules}, the LER performance progressively improves as we incrementally integrate the proposed modules.
Particularly, when the IDS decoder is added during the training stage, the LACC value improves significantly. This indicates that the IDS decoder enhances the extraction module by identifying whether characters contain complete radicals, thereby improving the performance of the character decoder through better character features.
For inference time, it can be observed that the time consumption is predominantly attributed to the localization module, primarily due to the processing of large-scale feature maps at this module.
When training exclusively with the handwritten training dataset and validating on the corresponding test dataset (all methods except SMTR in Table \ref{tab:ctr}), the LER is surpassed by that of other methods except CCR-CLIP. However, the CCR-CLIP \cite{yu2023chinese} excluded all non-Chinese samples from the handwritten scenario, resulting in a test dataset size of 11,018. In contrast, our experiment retained all test instances, yielding a larger test dataset of 23,389. Therefore, our approach achieved superior handwritten recognition results.

\begin{table}[t]
\centering
\caption{Effectiveness of char cutter depth.}
\begin{tabular}{ccc}
\hline
 $M$ & Handwriting & Parameters(M) \\
\hline
 1 & 58.40 / 0.862 & 21.8 \\
 3 & 59.29 / 0.871 & 25.2 \\
 6 & 59.32 / 0.865 & 30.7  \\
\hline
\end{tabular}%
\label{tab:ablation_depth}
\end{table}

\textbf{Effectiveness of char cutter depth.} We conducted a series of experiments to determine if increasing the number of char cutter iterations would enhance text recognition performance. According to Table \ref{tab:ablation_depth}, increasing the char cutter iterations $M$ from 1 to 3 significantly improves model accuracy. However, further increases beyond 3 iterations do not yield substantial improvements. Thus, the char cutter with $M=3$ represents the optimal balance between accuracy and model complexity.

\begin{table}[t]
\centering
\caption{Effectiveness of different model sizes. ``S'', ``B'' and ``L'' denote small, base and large, respectively. All experiments are conducted on the handwriting scenario of the CTR benchmark dataset.}
\resizebox{\columnwidth}{!}{%
\begin{tabular}{ccccc}
\hline
Models & {[}$D_0, D_1, D_2${]} & Heads & Params(M) & LACC / NED \\ \hline
LER-S & {[}96,192,256{]} & {[}3,6,8{]} & 16.4 & 57.31 / 0.858 \\
LER-B & {[}128,256,384{]} & {[}4,8,12{]} & 25.2 & 59.29 / 0.871 \\
LER-L & {[}192,256,512{]} & {[}6,8,16{]} & 32.1 & 60.34 / 0.886 \\ \hline
\end{tabular}%
}
\label{tab:ablation_modelsize}
\end{table}
\textbf{Effectiveness of different model sizes.}
We conducted a series of experiments to investigate whether increasing the model size could enhance the performance of LER. As shown in Table \ref{tab:ablation_modelsize}, we designed three model sizes: small, base, and large. In all three models, the image encoder consists of 6 conv mix blocks. In the localization module and IDS decoder, we use $N=6$, and in the extraction module, we use $M=3$. The results demonstrate that increasing the model parameters significantly improves LER performance. In this paper, when comparing our LER with other methods, we use the LER-B model by default.

\begin{table}[t]
\centering
\caption{Effectiveness of different character size.}
\begin{tabular}{ccc}
\hline
$c_h$ & $c_w$ & Handwriting \\
\hline
 3 & 3 & 59.22 / 0.864 \\
 4 & 3 & 58.87 / 0.861 \\
 4 & 4 & 59.29 / 0.871 \\
\hline
\end{tabular}%
\label{tab:ablation_charsize}
\end{table}
\textbf{Effectiveness of different character size.}
In the extraction module, selecting different character sizes may significantly impact recognition performance. If the character size is too small, it may fail to encompass all radicals of a character. Conversely, if the character size is too large, it may introduce overlapping features among decoupled characters. Therefore, we conducted a series of experiments to investigate the effects of varying character sizes on recognition accuracy, as shown in Table \ref{tab:ablation_charsize}. The results indicate that: 1. Given that Chinese characters are square-shaped, using square character size enhances recognition; 2. Since Chinese characters in text lines are typically written to fill the height of the line, a larger character size generally yields better performance.

\begin{figure*}[ht]
  \centering
  \includegraphics[width=1.0\linewidth]{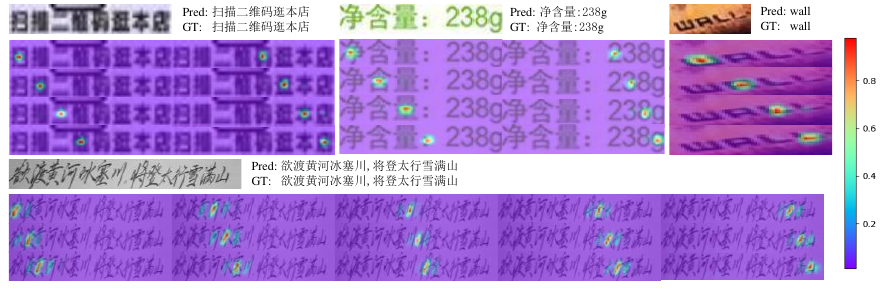}
  \caption{Visualization of the character localization.}
  \label{fig:visual_loc}
\end{figure*}

\begin{table}[t]
\centering
\caption{Effectiveness of training strategy.}
\begin{tabular}{cc}
\hline
Training strategy & Handwriting \\
\hline
 End-to-End & 56.36 / 0.852 \\
 Two stage  & 59.29 / 0.871 \\
\hline
\end{tabular}%
\label{tab:ablation_training}
\end{table}

\textbf{Effectiveness of training strategy.}
We design an experiment to validate the efficacy of the two-stage training strategy. As demonstrated in Table \ref{tab:ablation_training}, the two-stage strategy yields superior performance. This improvement arises because, during early end-to-end training, the localization module fails to provide precise character positions, which destabilizes the training of downstream modules. In contrast, the two-stage strategy guarantees robust localization capability before optimizing the extraction and recognition modules, thereby achieving better accuracy.

\begin{figure}[t]
  \centering
  \includegraphics[width=0.8\linewidth]{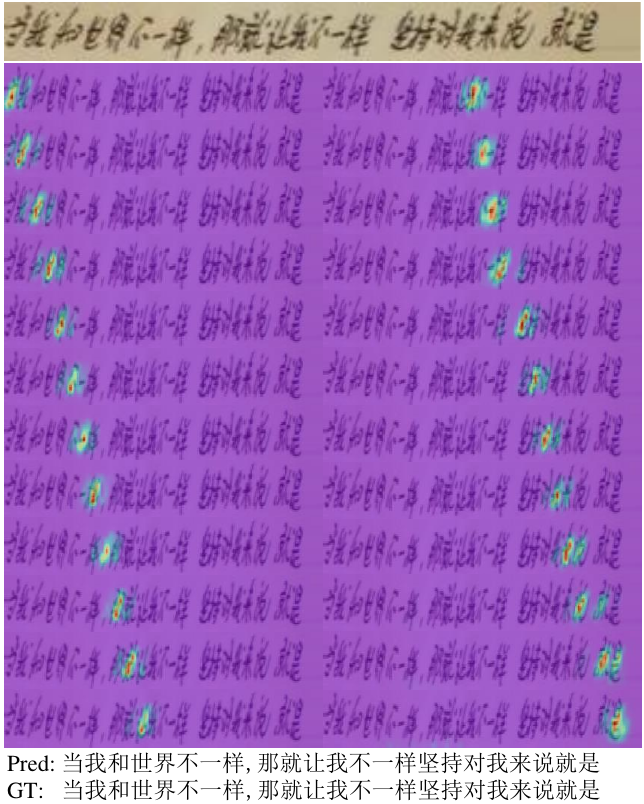}
  \caption{Visualization of the character localization on a long text instance.}
  \label{fig:visual_long}
\end{figure}

\subsection{Visual Analysis}
\textbf{Visualization of the character localization.}
In the LER network, the performance of the extraction module in obtaining decoupled character features hinges critically on the localization module, which supplies the char cutter with position-specific character features. Consequently, we visualize the character attention map from the localization module to investigate its localization accuracy, as illustrated in Fig. \ref{fig:visual_loc}. The visualization reveals that, in both English and Chinese text images, the localization module accurately locates characters at each position without attention drift, thus robustly supporting subsequent character decoupling and recognition.

In Chinese handwritten text images, long text sequences are frequently encountered. To investigate the recognition performance of LER on long text images and to determine whether attention drift occurs, we provide a visualization of a long text image instance, as illustrated in Fig. \ref{fig:visual_long}.

\begin{figure}[t]
  \centering
   \includegraphics[width=1.0\linewidth]{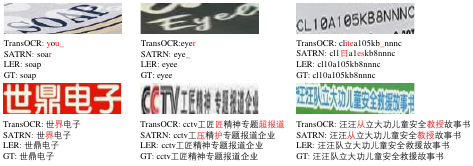}
   \caption{Visualization of the predictions from different methods. Red character denotes the error prediction.}
   \label{fig:visual_compare}
\end{figure}

\textbf{Visualization of the predictions from different methods.}
We present the predictions of different methods on the Web scenario of the CTR benchmark dataset to validate the effectiveness of the proposed LER. All methods are trained exclusively on the Web training dataset and validated on the Web test dataset. Although the CTR benchmark primarily targets Chinese text recognition, it also includes a substantial number of English instances. As shown in Fig. \ref{fig:visual_compare}, the first row illustrates the recognition comparison for English text, while the second row represents the comparison for Chinese text. Both TransOCR and SATRN employ AR decoding methods without language model post-processing. In the first two images of the first row, due to the AR models learning lexicon dependencies during training, they tend to recognize similar meaningful words during inference, such as SATRN falsely recognizes ``soap'' as ``soar'' and ``eyee'' as ``eye". When dealing with meaningless text images, the LER model demonstrates a significant advantage. The third image in the first row shows a meaningless text image, where methods based on AR decoder are prone to multiple recognition errors, whereas the LER model avoids such issues.

\subsection{Error Analysis}
\begin{figure}[t]
  \centering
   \includegraphics[width=1.0\linewidth]{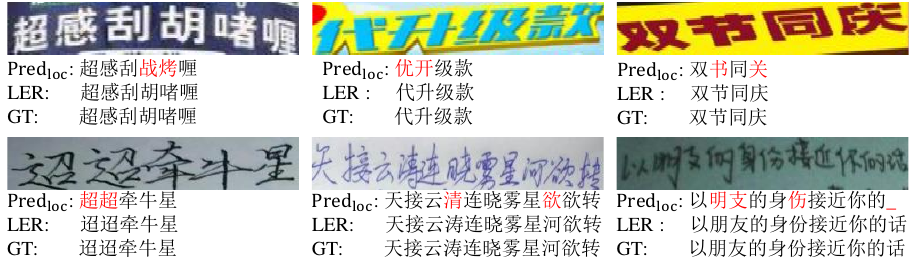}
   \caption{The text images with errors predicted by the localization module are correctly recognized by the complete LER model. Red character denotes the error prediction.}
   \label{fig:errorLocLER}
\end{figure}

In the LER model, the localization module is inherently an AR decoder and thus provides a prediction for the text. However, once a character error occurs, subsequent recognition becomes unreliable, potentially leading to multiple recognition errors, as illustrated in Fig. \ref{fig:errorLocLER}. In contrast, the complete LER model, which employs a character decoupling approach for text prediction, is not influenced by previously predicted incorrect characters and can achieve more accurate recognition results.
This advantage is further supported by Table \ref{tab:ablation_modules}, which shows that the complete LER model achieves a 2.07\% improvement in accuracy compared to using only the localization module for prediction.

\begin{figure}[t]
  \centering
   \includegraphics[width=1.0\linewidth]{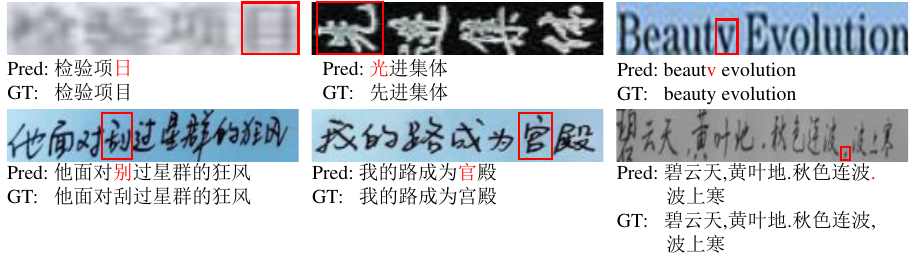}
   \caption{Error predictions from LER, the red box highlights falsely recognized characters, while red text denotes incorrect predictions.}
   \label{fig:errorLER}
\end{figure}

Although LER achieves relatively excellent recognition performance, it may still fail in some challenging instances. Fig. \ref{fig:errorLER} illustrates several cases where LER encounters failures. The following situations pose significant challenges to the LER model: 
1) Poor image quality. Low-quality images present challenges for both humans and recognition methods, as shown in the first image of the first row.
2) Vertical text images. Despite rotating images with an aspect ratio greater than 2 by 90 degrees during training and testing to make them a horizontal version, rotated text remains challenging for character recognition, especially when characters are similar to existing ones.
3) Errors caused by damaged characters or similar-shape characters. Since LER employs a character decoupling approach for recognition, it is susceptible to error when characters in the image are damaged or resemble other characters. For instance, in the text image labeled ``Beauty Evolution,'' the ``y'' in ``Beauty'' is incorrectly predicted as ``v'' because ``y'' is damaged. Furthermore, in handwritten images, diverse writing styles increase the risk of error due to similar-shaped characters.

\begin{figure}[t]
  \centering
   \includegraphics[width=1.0\linewidth]{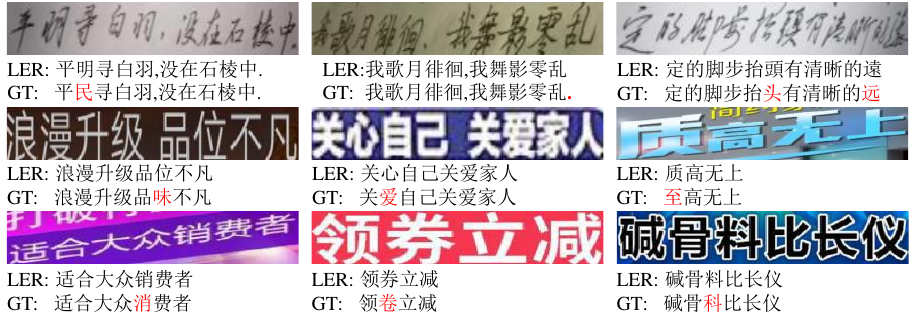}
   \caption{Some error labels in the CTR benchmark result in LER's correct predictions being marked as errors. Red character denotes incorrect labels.}
   \label{fig:errorLabel}
\end{figure}

Interestingly, we discovered instances of incorrect labeling in the CTR benchmark dataset. In some cases, annotators may have introduced subjective linguistic inferences, resulting in labels that do not accurately reflect the text in the images. 
In such cases, LER, which employs a character decoupling approach and independently recognizes each character, correctly identified these images despite the labeling errors. Nevertheless, due to the wrong labels, LER's accurate predictions are incorrectly classified as failures. Fig. \ref{fig:errorLabel} illustrates several images where LER accurately recognizes the text despite incorrect labeling. 
In these instances, annotators may have inadvertently replaced the characters in the image with homophones, as an example from the first image of the second row; or they may have inadvertently modified the wrongly written characters in the image, as an example from the first image of the third row.
It is evident that LER faithfully follows the original characters in the images, whereas human annotators may introduce linguistic interference, resulting in labeling errors. The AR-based decoder model and the LM post-processing model adopt a similar approach to human annotators when recognizing characters, often introducing lexicon dependencies. 
However, in specific scenarios, automatically correcting erroneous characters in images is not permissible. For example, if an examination paper contains misspelled words and an OCR model is used to recognize the text, proactively correcting these misspellings based on lexicon information would distort the assessment of the examinee's performance. Accurate evaluation requires that the original errors be preserved to reflect the true nature of the examinee's responses.

\section{Conclusion}
In this paper, we present LER, a Chinese text recognition method that avoids error accumulation and faithfully recognizes characters in images without automatic error correction. LER presents a novel character-level decoupled text recognition approach. The LER network independently recognizes characters through three modules. First, the multimodal localization module identifies the position of each character within the image. Then, the feature extraction module isolates and extracts features for each character, obtaining relatively independent character features. Finally, the recognition module provides the recognition result for each character.

To enhance the model's performance in Chinese text recognition, we introduce an IDS decoder during the training stage. This decoder can be removed during inference to reduce model size and improve inference efficiency. Additionally, we deploy a two-stage training strategy to ensure more effective training. Experimental results demonstrate that LER achieves state-of-the-art performance on Chinese benchmarks and delivers competitive results on English benchmarks, even with a smaller model size. However, the number of features output by the localization module is equal to the supported maximum number of characters, resulting in significant redundancy and negatively impacting the model's speed. Therefore, our future work will focus on addressing this issue to optimize the model's efficiency further.

 {
\bibliographystyle{IEEEtran}
\bibliography{IEEEabrv, bib}
}


 
\vspace{11pt}

\begin{IEEEbiography}[{\includegraphics[width=1in,height=1.25in,clip,keepaspectratio]{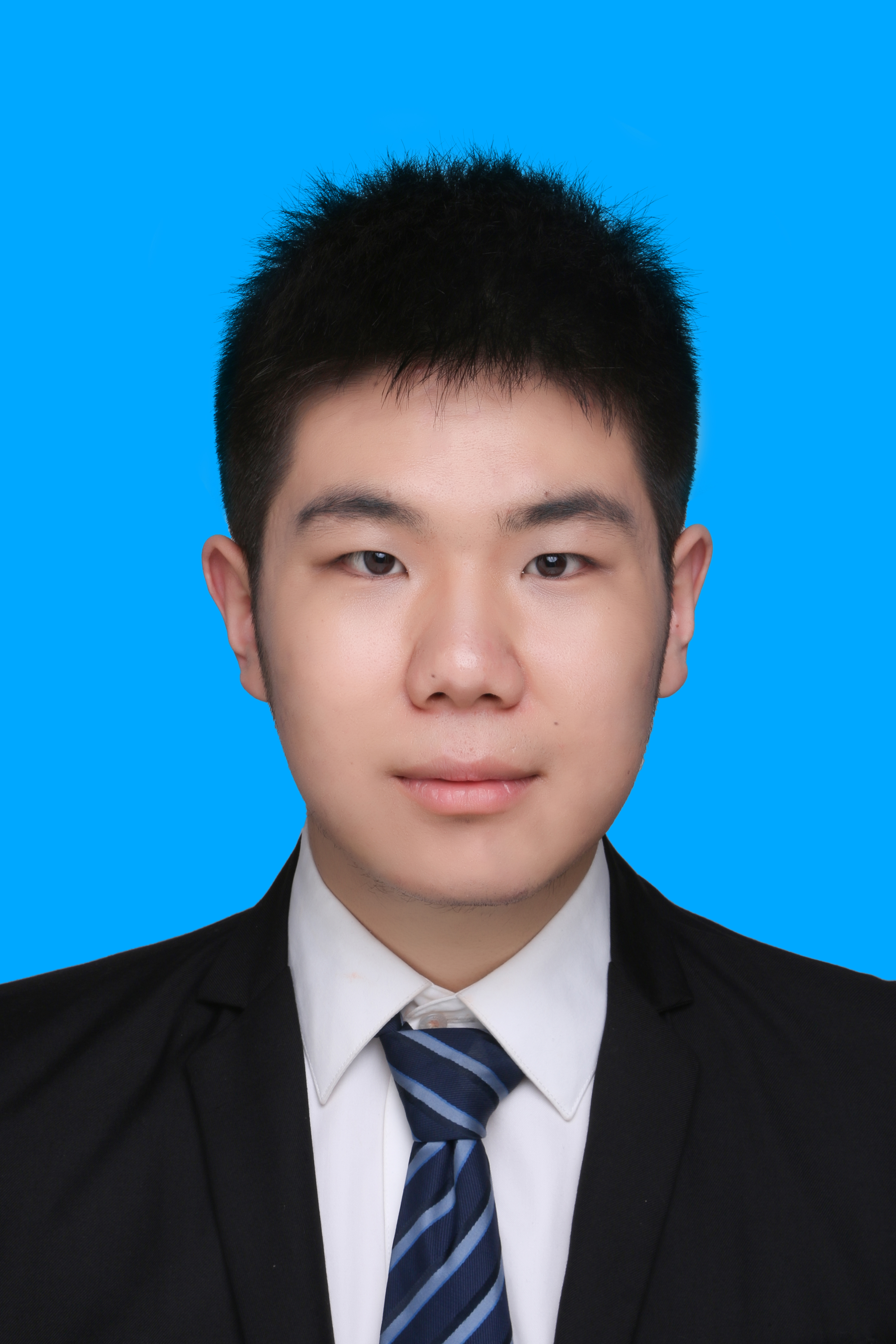}}]{Qilong Li} received the M.S. degree in the school of Electrical Engineering at the Yanshan University, Qin Huangdao, China in 2021. He is currently pursuing the Ph.D. degree in the School of Computer and Information Engineering, Henan University. His research interests include Chinese text/character recognition and computer vision.
\end{IEEEbiography}

\vspace{11pt}

\begin{IEEEbiography}[{\includegraphics[width=1in,height=1.25in,clip,keepaspectratio]{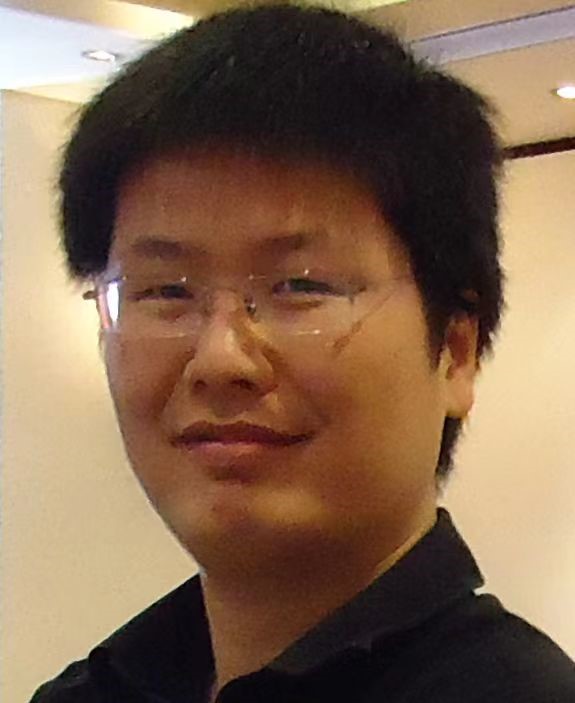}}]
{Chongsheng Zhang}
(M'20-SM'23) obtained his Ph.D. degree from INRIA, France. Since then, he has been a full professor of Henan University, China, where he leads the data science and artificial intelligence (DSAI) research team.  His research interests include imbalance learning, OCR and digital paleography. He has published 50 papers in peer-reviewed journals and conferences, including SIGKDD 2022, AAAI 2023, IJCAI 2020, etc. He has authored 6 books and filed 20 Chinese patents.
\end{IEEEbiography}

\vfill

\end{document}